\documentclass[runningheads]{llncs}
\usepackage{graphicx}
\usepackage{amsmath,amssymb} 
\usepackage{color}
\usepackage[width=122mm,left=12mm,paperwidth=146mm,height=193mm,top=12mm,paperheight=217mm]{geometry}
\usepackage{subfigure}
\usepackage{bm}
\usepackage{algpseudocode}
\usepackage{caption}
\usepackage{algorithm}
\usepackage{multicol}
\usepackage{multirow}
\usepackage{subfigure}
\begin{document}
\pagestyle{headings}
\mainmatter

\title{Learning Class Prototypes via Structure Alignment for Zero-Shot Recognition}

\titlerunning{Learning Class Prototypes via Structure Alignment for ZSL}

\authorrunning{H. Jiang, R. Wang, S. Shan, X. Chen}

\author{Huajie Jiang\inst{1,2,3,4}, Ruiping Wang\inst{1,4}, Shiguang Shan\inst{1,4}, Xilin Chen\inst{1,4}}


\institute{$^1$Key Laboratory of Intelligent Information Processing of Chinese Academy of Sciences (CAS),
Institute of Computing Technology, CAS, Beijing, 100190, China\\
$^2$Shanghai Institute of Microsystem and Information Technology, CAS, Shanghai, 200050, China\\
$^3$ShanghaiTech University, Shanghai, 200031, China\\
$^4$University of Chinese Academy of Sciences, Beijing, 100049, China\\
\email{ huajie.jiang@vipl.ict.ac.cn, \{wangruiping, sgshan, xlchen\}@ict.ac.cn}
}

\maketitle

\begin{abstract}
Zero-shot learning (\textbf{ZSL}) aims to recognize objects of novel classes without any training samples of specific classes, which is achieved by exploiting the semantic information and auxiliary datasets. Recently most \textbf{ZSL} approaches focus on learning visual-semantic embeddings to transfer knowledge from the auxiliary datasets to the novel classes. However, few works study whether the semantic information is discriminative or not for the recognition task. To tackle such problem, we propose a coupled dictionary learning approach to align the visual-semantic structures using the class prototypes, where the discriminative information lying in the visual space is utilized to improve the less discriminative semantic space. Then, zero-shot recognition can be performed in different spaces by the simple nearest neighbor approach using the learned class prototypes. Extensive experiments on four benchmark datasets show the effectiveness of the proposed approach.
\end{abstract}
\section{Introduction}

Object recognition has made tremendous progress in recent years. With the emergence of large-scale image database \cite{ILSVRC15}, deep learning approaches \cite{Krizhevsky2012,Christian2015,Simonyan2014,Kaiming2016} show their great power to recognize objects. However, such supervised learning approaches require large numbers of images to train robust recognition models and can only recognize a fixed number of categories, which limits their flexibility. It is well known that collecting large numbers of images is difficult. On one hand, the numbers of images often follow a long-tailed distribution \cite{Zhu2014CapturingLD} and it is hard to collect images for some rare categories. On the other hand, some fine-grained annotations require expert knowledge \cite{Wah2011TheCB}, which increases the difficulty of the annotation task. All these challenges motivate the rise of zero-shot learning, where no labeled examples are needed to recognize one category.

Zero-shot learning aims at recognizing objects that have not been seen in the training stage, where auxiliary datasets and semantic information are needed to perform such tasks. It is mainly inspired by the human's behavior to recognize new objects. For example, children have no problem recognizing \emph{zebra} if they are told that \emph{zebra} looks like a \emph{horse} (auxiliary datasets) but has \emph{stripes} (semantic information), even though they have never seen \emph{zebra} before. Current \textbf{ZSL} approaches generally involve three steps. First, choose a semantic space to build up the relations between seen (auxiliary dataset) and unseen (test) classes.  Recently the most popular semantic information includes attributes \cite{lampert2009learning,farhadi2009describing} that are manually defined and wordvectors \cite{frome2013devise,akata2015evaluation} that are automatically extracted from the auxiliary text corpus. Second, learn general visual-semantic embeddings from the auxiliary dataset, where the images and class semantics could be projected into a common space \cite{akata2013label,Changpinyo2017PredictingVE}. Third, perform the recognition task in the common space by different metric learning approaches.

\begin{figure}[t]
\centering
\includegraphics[height=4.2cm]{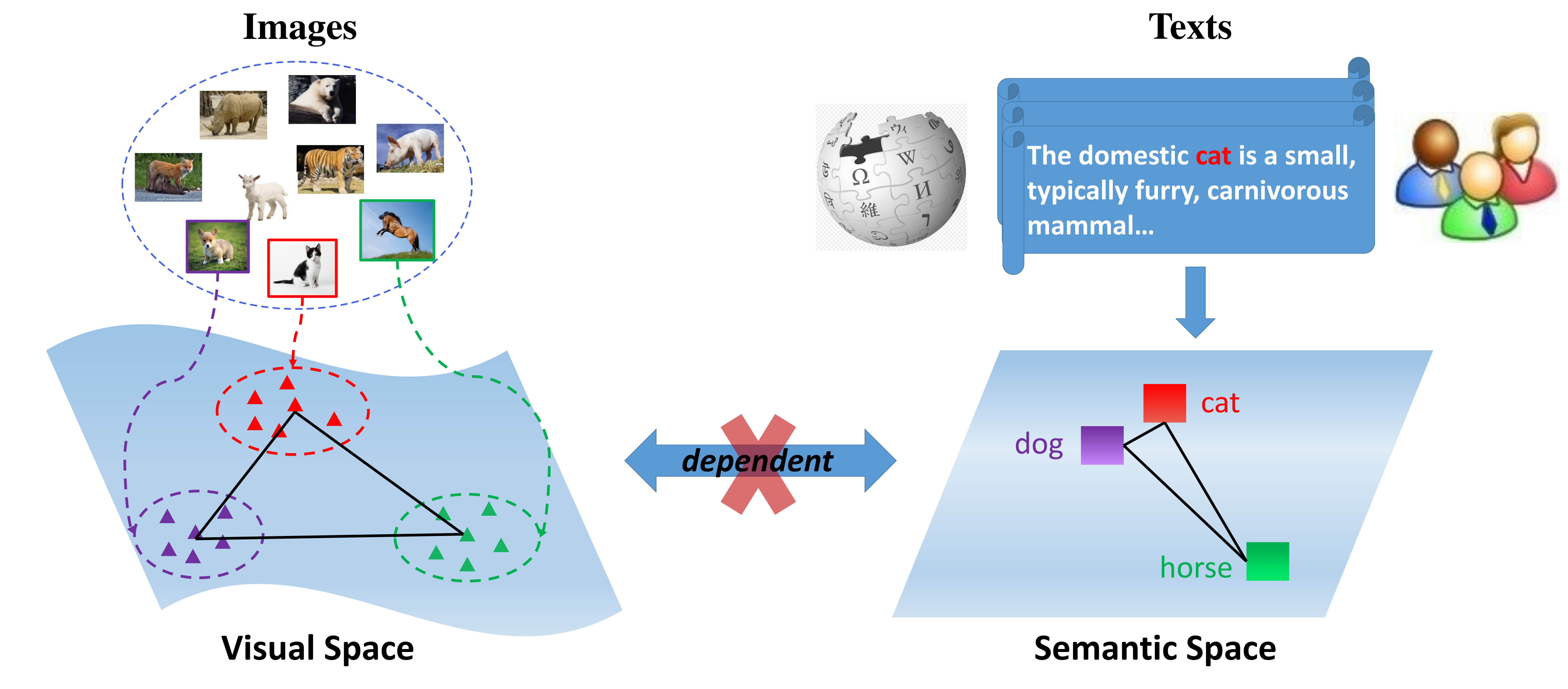}
\caption{Illustration diagram that shows the inconsistency of visual feature space and semantic space. The semantic information is manually defined or automatically extracted, which is independent of visual samples. The black lines in the two spaces show the similarities between different classes.}
\label{fig:structure}
\end{figure}

Traditional \textbf{ZSL} approaches usually use fixed semantic information and pay much attention to learning more robust visual-semantic embeddings \cite{lampert2009learning,frome2013devise,akata2013label,Kodirov2015UnsupervisedDA,romera2015embarrassingly,Zhang2017LearningAD}. However, most of these approaches ignore the fact that the semantic information, whether human-defined or automatically extracted, is incomplete and may be not discriminative enough to classify different classes because the descriptions about classes are limited. As is shown in Figure \ref{fig:structure}, some classes may locate quite close to each other in the semantic space due to the incomplete descriptions, \emph{i.e.} \emph{cat} and \emph{dog}, thus it may be less effective to perform recognition task in this space. Since images are real reflections of different categories, they may contain more discriminative information that could not be described. Moreover, the semantic information is obtained independently from visual samples so the class structures between the visual space and semantic space are not consistent. In such cases, the visual-semantic embeddings would be too complicated to learn. Even if the embeddings are properly learned, they have large probabilities to overfit the seen classes and have less expansibility to the unseen classes.

In order to tackle such problems, we propose to learn the class prototypes by aligning the visual-semantic structures. The novelty of our framework lies in three aspects. First, different from traditional approaches which learn image embeddings, we perform the structure alignment on the class prototypes, which are automatically learned, to conduct the recognition task. Second, a coupled dictionary learning framework is proposed to align the class structures between visual space and semantic space, where the discriminative property lying in the visual space and the extensive property existing in the semantic space are merged in an aligned space. Third, semantic information of unseen classes is utilized for domain adaptation, which increases the expansibility of our model to the unseen classes. In order to demonstrate the effectiveness of the proposed approach, we perform experiments on four popular datasets for zero-shot recognition, where excellent results are achieved.


\section{Related Work}

In this section, we review related works on zero-shot learning in three aspects, \emph{i.e.} semantic information, visual-semantic embeddings, zero-shot recognition.

\subsection{Semantic Information}

Semantic information plays an important role in zero-shot learning. It builds up the relations between seen and unseen classes, thus making it possible for zero-shot recognition. Recently, the most popular semantic information includes attributes \cite{lampert2009learning,farhadi2009describing,akata2013label,Bucher2016ImprovingSE,Jiang2017LearningDL} and wordvectors \cite{akata2015evaluation,Demirel2017Attributes2ClassnameAD,Morgado2017SemanticallyCR}. Attributes are general descriptions of objects which can be shared among different classes. For example, {\it furry} can be shared among different animals. Thus it is possible to learn such attributes by some auxiliary classes and apply them to the novel classes for recognition. Wordvectors are automatically extracted from large numbers of text corpus, where the distances between different wordvectors show the relations between different classes, thus they are also capable of building up the relations between seen and unseen classes.

Since the knowledge that could be collected is limited, the semantic information obtained in general purpose is usually less discriminative to classify different classes in specific domains. To tackle such problem, we propose to utilize the discriminative information lying in the visual space to improve the semantic space.

\subsection{Visual-Semantic Embeddings}

Visual-semantic embedding is the key to zero-shot learning and most existing \textbf{ZSL} approaches focus on learning more robust visual-semantic embeddings. In the early stage, \cite{lampert2009learning,farhadi2009describing} propose to use attribute classifiers to perform \textbf{ZSL} task. Such methods learn each attribute classifier independently, which is not applicable to large-scale datasets with lots of attributes. In order to tackle such problems, label embedding approaches emerge \cite{akata2013label,akata2015evaluation}, where all attributes are considered as a whole for a class and label embedding functions are learned to maximize the compatibility of images with corresponding class semantics. To improve the performance of such embedding models, \cite{Xian2016LatentEF} proposes latent embedding models, where multiple linear embeddings are learned to approximate non-linear embeddings. Furthermore, \cite{frome2013devise,Socher2013ZeroShotLT,Wang2016LearningD,Reed2016LearningDR,Zhang2017LearningAD,Morgado2017SemanticallyCR} exploit deep neural networks to learn more robust visual-semantic transformations.

Although some works pay attention to learning more complicated embedding functions, some other works deal with the visual-semantic transformation problem from different views. \cite{Norouzi2014ZeroShotLB} forms the semantic information of unseen samples by a convex combination of seen-class semantics. \cite{Zhang2015ZeroShotLV,Zhang2016ZeroShotLV} utilize the class similarities and \cite{Jiang2017LearningDL} proposes discriminative latent attributes to form more effective embedding space. \cite{Changpinyo2016SynthesizedCF} synthesizes the unseen-class classifiers by sharing the structures between the semantic space and the visual space. \cite{Changpinyo2017PredictingVE,Long2017ZeroshotLU} predicts the visual exemplars by learning embedding functions from the semantic space to the visual space. \cite{Bucher2016ImprovingSE} exploits metric learning techniques, where relative distance is utilized, to improve the embedding models. \cite{RomeraParedes2015AnES} views the image classifier as a function of corresponding class semantic and uses additional regularizer to learn the embedding functions. \cite{Kodirov2017SemanticAF} utilizes the auto-encoder framework to learn the visual-semantic embeddings. \cite{Ding2017LowRankEE} uses low rank constraints to learn semantic dictionaries and \cite{Xu2017MatrixTW} proposes a matrix tri-factorization approach with manifold regularizations. To tackle the embedding domain shift problem, \cite{Kodirov2015UnsupervisedDA,Fu2015TransductiveMZ} use the transfer learning techniques to extend \textbf{ZSL} into transductive settings, where the unseen-class samples are also utilized in the training process.

Different from such existing approaches which learn image embeddings or synthesize image classifiers, we propose to learn the class prototypes by jointly aligning the class structures between the visual space and the semantic space.

\subsection{Zero-Shot Recognition}

The most widely used approaches for zero-shot recognition are probability models \cite{lampert2009learning} and nearest neighbour classifiers \cite{akata2013label,Zhang2015ZeroShotLV,Jiang2017LearningDL}. To make use of the rich intrinsic structures on the semantic manifold, \cite{Fu2015ZeroshotOR} proposes semantic manifold distance to recognize the unseen class samples and \cite{Changpinyo2016SynthesizedCF} directly synthesizes the image classifiers of unseen classes in the visual space by sharing the structures between the semantic space and the visual space. Considering more real conditions, \cite{Chao2016AnES} expands the traditional \textbf{ZSL} problem to the generalized \textbf{ZSL} problem, where the seen classes are also considered in the test procedure. Recently, \cite{Xian2017ZeroShotL} proposes more reasonable data splits for different datasets and evaluates the performance of different approaches under such experiment settings.

\section{Approaches}

\begin{figure}[t]
\centering
\includegraphics[height=4.5cm]{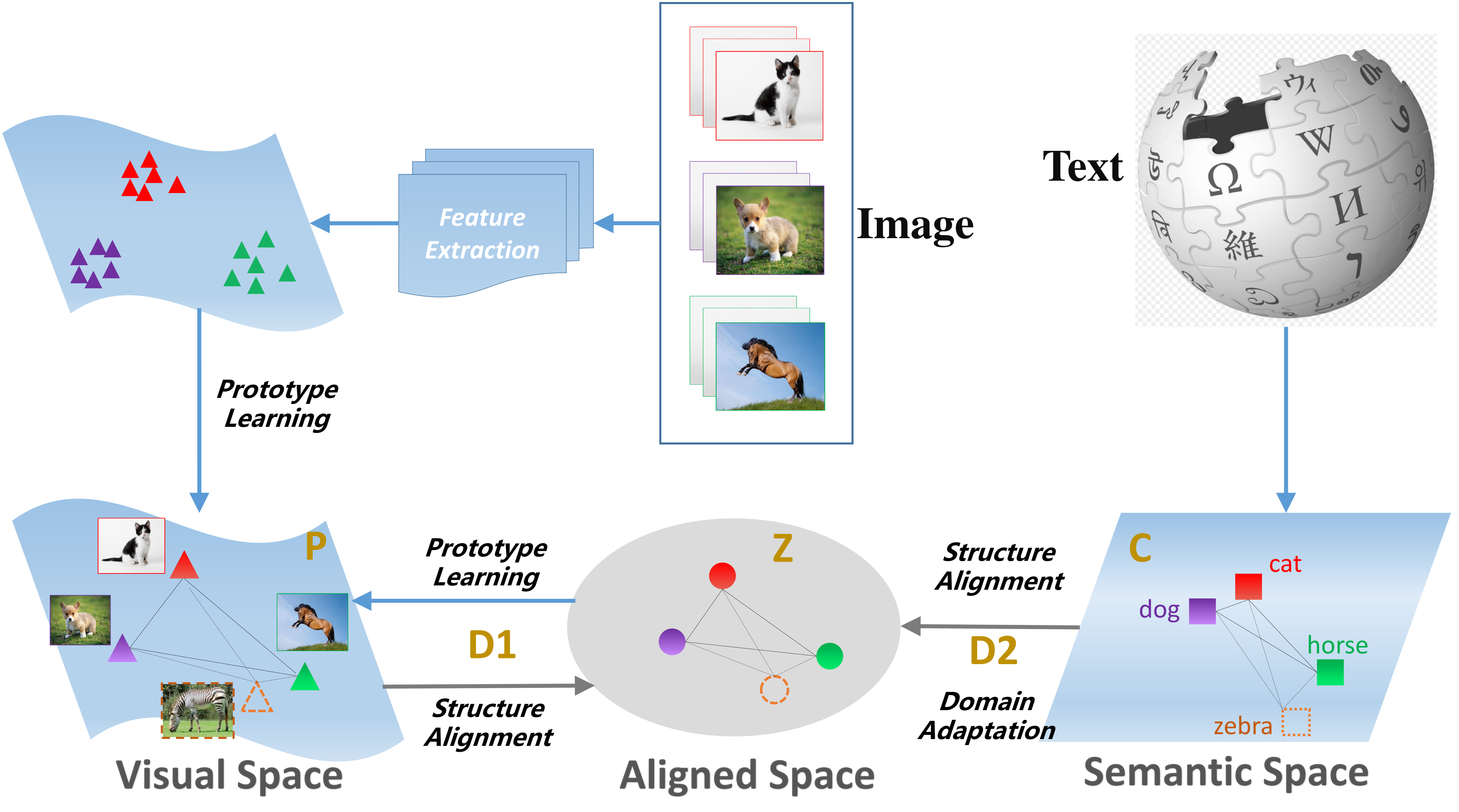}
\caption{Coupled dictionary learning framework to align the visual-semantic structure. The solid shapes represent the seen-class prototypes and the dotted shapes denote the prototypes of unseen classes. Black lines show the relationships between different classes. The brown characters are corresponding to the formulation of equations.}
\label{fig:framework}
\end{figure}

The general idea of the proposed approach is to learn the class prototypes by sharing the structures between the visual space and the semantic space. However, the structures between these two spaces may be inconsistent, since the semantic information is obtained independently of the visual examples. In order to tackle such problem, we propose a coupled dictionary learning (\textbf{CDL}) framework to simultaneously align the visual-semantic structures. Thus the discriminative information in the visual space and the relations in the semantic space can be shared to benefit each other. Figure \ref{fig:framework} shows the framework of our approach. There are three key submodules of the proposed framework: prototype learning, structure alignment, and domain adaptation.

\subsection{Problem Formulation}

Assume a labeled training dataset contains $K$ seen classes with $n_s$ labeled samples $ \mathcal{S} = \{(x_i, y_i)| x_i \in \mathcal{X}, y_i \in \mathcal{Y}^{s}\}_{i=1}^{n_s}$, where $x_i \in \mathbb{R}^d $ represents the image feature and $y_i$ denotes the class label in $\mathcal{Y}^{s}=\{s_1,...,s_K\}$. In addition, a disjoint class label set $\mathcal{Y}^{u}=\{u_1,...,u_L\}$, which consists $L$ unseen classes, is provided, \emph{i.e.} $\mathcal{Y}^{u} \bigcap \mathcal{Y}^{s} = \O $, but the corresponding images are missing. Given the class semantics $\mathcal{C} = \{\mathcal{C}^s \bigcup \mathcal{C}^u\}$, the goal of \textbf{ZSL} is to learn image classifiers $f_{zsl} : \mathcal{X} \rightarrow \mathcal{Y}^{u}$.

\subsection{Framework}

As is shown in Figure \ref{fig:framework}, our framework contains three submodules: prototype learning, structure alignment and domain adaptation.

\textbf{Prototype Learning} \quad The structure alignment approach proposed by our framework is performed on the class prototypes. In order to align the class structures between the visual space and the semantic space, we must first obtain the class prototypes in both spaces. In the semantic space, we denote the class prototypes of seen/unseen classes as $C_s \in \mathbb{R}^{m \times K}$/$C_u \in \mathbb{R}^{m \times L}$, where $m$ is the dimension of the semantic space. Here, $C_s$/$C_u$ can be directly set as $\mathcal{C}^s$/$\mathcal{C}^u$. However, in the visual space, only the seen-class samples $X_s \in \mathbb{R}^{d \times n_s}$ and their corresponding labels $Y_s$ are provided, so we should first learn the class prototypes $P_s \in \mathbb{R}^{d \times K}$ in the visual space, where $d$ is the dimension of the visual space. The basic idea for prototype learning is that samples should locate near their corresponding class prototypes in the visual space, so the loss function can be formulated as:
\begin{equation}
  \mathcal{L}_p = \min \limits_{P_s} \left \| X_s - P_sH \right \|_{F}^{2}, \\ \quad
  \label{equ:proto}
\end{equation}
where each column in $H \in \mathbb{R}^{K \times n_s}$ is a one-hot vector indicating the class label of corresponding image.

\textbf{Structure Alignment} \quad Due to the fact that the semantic information of classes is defined or extracted independently of the images, directly sharing the structures in the semantic space to form the prototypes of unseen classes in the visual space is not a good choice, where structure alignment should be performed first. Therefore, we propose a coupled dictionary learning framework to align the visual-semantic structures. The basic idea for our structure alignment approach is to find some bases in each space to represent each class and enforce the new representation to be the same in the two spaces, thus the structures can be aligned. The loss function is formulated as:
\begin{equation}
\begin{split}
  \mathcal{L}_s = & \min \limits_{P_s,D_1,D_2,Z_s} \left \| P_s - D_1Z_s \right \|_{F}^{2} + \lambda \left \| C_s - D_2Z_s \right \|_{F}^{2}, \\
  & s.t. \quad ||\bm{d}_1^i||_2^2 \leq 1, \quad ||\bm{d}_2^i||_2^2 \leq 1, \forall i.
  \label{equ:struct}
\end{split}
\end{equation}
where $P_s$ and $C_s$ are the prototypes of seen classes in the visual and semantic space respectively. $D_1 \in \mathbb{R}^{d \times n_b}$ and $D_2 \in \mathbb{R}^{m \times n_b}$ are the bases in corresponding spaces, where $d,m$ are the dimensions of visual space and semantic space respectively and $n_b$ is the number of bases. $Z_s \in \mathbb{R}^{n_b \times K}$ is the common new representation of seen classes, and it just plays the key role to align the two spaces. $\lambda$ is a parameter controlling the relative importance of the visual space and semantic space. $\bm{d}_1^i$ denotes the $i$-th column of $D_1$ and $\bm{d}_2^i$ is the $i$-th column of $D_2$. By exploring new representation bases in each space to reformulate each class, we obtain the same class representations for the visual and semantic spaces, thus the class structures in the two spaces will be consistent.

\textbf{Domain Adaptation} \quad In the structure alignment process, only seen-class prototypes are utilized and this may cause the domain shift problem \cite{Fu2015TransductiveMZ}. In other words, a general structure alignment approach learned on seen classes may not be appropriate for the unseen classes, since there are some differences between seen and unseen classes. To tackle such problem, we further propose a domain adaptation term, which automatically learns the unseen-class prototypes in the visual space and uses the unseen prototypes to assist the structure learning process. The loss function can be formulated as:
\begin{equation}
\begin{split}
  \mathcal{L}_u = & \min \limits_{P_u,D_1,D_2,Z_u} \left \| P_u - D_1Z_u \right \|_{F}^{2} + \lambda \left \| C_u - D_2Z_u \right \|_{F}^{2}, \\
  & s.t. \quad ||\bm{d}_1^i||_2^2 \leq 1, \quad ||\bm{d}_2^i||_2^2 \leq 1, \forall i.
  \label{equ:adapt}
\end{split}
\end{equation}
where $P_u \in \mathbb{R}^{d \times L}$ and $C_u \in \mathbb{R}^{m \times L}$ are the prototypes of unseen classes in the visual and semantic space respectively, and $Z_u \in \mathbb{R}^{n_b \times L}$ is the common new representation of unseen classes.

In a whole, our full objective can be formulated as:
\begin{equation}
\begin{split}
  \mathcal{L} = \mathcal{L}_s + \alpha \mathcal{L}_u + \beta \mathcal{L}_p,
  \label{equ:whole}
\end{split}
\end{equation}
where $\alpha$ and $\beta$ are the parameters controlling the relative importance.

\subsection{Optimization}

The final loss function of the proposed framework can be formulated as:
\begin{equation}
\begin{split}
  \mathcal{L} = & \min \limits_{P_s,P_u,D_1,D_2,Z_s,Z_u} (\left \| P_s - D_1Z_s \right \|_{F}^{2} + \lambda \left \| C_s - D_2Z_s \right \|_{F}^{2}) + \\
  & \alpha (\left \| P_u - D_1Z_u \right \|_{F}^{2} + \lambda \left \| C_u - D_2Z_u \right \|_{F}^{2}) + \beta(\left \| X_s - P_sH \right \|_{F}^{2}),\\
  & s.t. \quad ||\bm{d}_1^i||_2^2 \leq 1, \quad ||\bm{d}_2^i||_2^2 \leq 1, \forall i.
  \label{equ:whole2}
\end{split}
\end{equation}
It is obvious that Eq.\ref{equ:whole2} is not convex for $P_s, P_u, D_1, D_2, Z_s$ and $Z_u$ simultaneously, but it is convex for each of them separately. We thus employ an alternating optimization method to solve the problem.

\textbf{Initialization} \quad In our framework, we set the number of dictionary bases $n_b$ as the number of seen classes $K$ and enforces each column of $Z$ to be the similarities to all seen classes. First, we initialize $Z_u \in \mathbb{R}^{K \times L}$ as the similarities of unseen classes to the seen classes, \emph{i.e.} cosine distances between unseen and seen class prototypes in the semantic space. Second, we get $D_2$ by the second term of Eq.\ref{equ:adapt}, which has closed-form solution. Third, we get $Z_s$ by the second term of Eq.\ref{equ:struct}. Next, we initialize $P_s$ as the mean of samples in each class. Then, we get $D_1$ by the first term of Eq.\ref{equ:struct}. In the end, we get $P_u$ by the first term in Eq.\ref{equ:adapt}. In this way, all the variables in our framework are initialized.

\textbf{Joint Optimization} After all variables in our framework are initialized separately, we jointly optimize them as follows:

(1) Fix $D_1,Z_s$ and update $P_s$. The subproblem can be formulated as:
\begin{equation}
\begin{split}
  \arg \min \limits_{P_s}  \left \| P_s - D_1Z_s \right \|_{F}^{2}  +  \beta \left \| X_s - P_sH \right \|_{F}^{2}
  \label{equ:sub1}
\end{split}
\end{equation}

(2) Fix $P_s,D_1,D_2$ and update $Z_s$ by Eq.\ref{equ:struct}.

(3) Fix $P_s,P_u,Z_s,Z_u$ and update $D_1$. The subproblem can be formulated as:
\begin{equation}
\begin{split}
  \arg \min \limits_{D_1}  \left \| P_s - D_1Z_s \right \|_{F}^{2}  + \alpha \left \| P_u - D_1Z_u \right \|_{F}^{2} \quad  s.t. \quad ||\bm{d}_1^i||_2^2 \leq 1, \forall i.
  \label{equ:sub3}
\end{split}
\end{equation}

(4) Fix $Z_s,Z_u$ and update $D_2$. The subproblem can be formulated as:
\begin{equation}
\begin{split}
  \arg \min \limits_{D_1}  \left \| C_s - D_2Z_s \right \|_{F}^{2}  + \alpha \left \| C_u - D_2Z_u \right \|_{F}^{2} \quad  s.t. \quad ||\bm{d}_2^i||_2^2 \leq 1, \forall i.
  \label{equ:sub4}
\end{split}
\end{equation}

(5) Fix $P_u,D_1,D_2$ and update $Z_u$ by Eq.\ref{equ:adapt}.

(6) Fix $D_1,Z_u$ and update $P_u$ by the first term of Eq.\ref{equ:adapt}.

In our experiments, we set the maximum iterations as 100 and the optimization always converges after tens of iterations, usually less than 50. \footnote{Source code of CDL is available at \emph{http://vipl.ict.ac.cn/resources/codes}.}

\subsection{Zero-Shot Recognition}
\label{sec:recognition}

In the proposed framework, we can obtain the prototypes of unseen classes in different spaces (\emph{i.e.} visual space $P_u$, aligned space $Z_u$, semantic space $C_u$), where we can perform zero-shot recognition task using nearest neighbour approach.

\textbf{Recognition in the Visual Space}. \quad In the test process, we can directly compute the similarities $Sim_v$ of test samples ($X_i$) to the unseen class prototypes ($P_u$), \emph{i.e.} cosine distance, and classify the images to the classes corresponding to their most similar prototypes.

\textbf{Recognition in the Aligned Space}. \quad To perform recognition task in this space, we must first obtain the representations of images in this space by
\begin{equation}
\begin{split}
  \arg \min \limits_{Z_i}  \left \| X_i - D_1Z_i \right \|_{F}^{2} + \gamma \left \|Z_i \right \|_{F}^{2}
  \label{equ:testl}
\end{split}
\end{equation}
where $X_i$ represents the test images and $Z_i$ is the corresponding representation in the aligned space. Then we can obtain the similarities $Sim_a$ of test samples ($Z_i$) to the unseen-class prototypes ($Z_u$) and use the same recognition approach as that in the visual space.

\textbf{Recognition in the Semantic Space}. \quad First, we should get the semantic representations of images by $C_i = D_2Z_i$. Then the similarities $Sim_s$ can be obtained by computing the distances between the test samples ($C_i$) and the unseen-class prototypes ($C_u$). The recognition task can be performed the same way as that in the visual space.

\textbf{Combining Multiple Spaces}. \quad Due to the fact that the visual space is discriminative, the semantic space is more generative, and the aligned space is a compromise, combining multiple spaces would improve the performance. In our framework, we simply combine the similarities obtained in each space, \emph{i.e.} combining the visual space and aligned space by $Sim_{va} = Sim_v + Sim_a$, and use the same nearest neighbour approach to perform recognition task.

\subsection{Difference from Relevant Works}

Among prior works, the most relevant one to ours is \cite{Changpinyo2016SynthesizedCF}, where the structures in the semantic space and visual space are also utilized. However, the key ideas of the two works are quite different. \cite{Changpinyo2016SynthesizedCF} uses fixed semantic information and directly shares its structure to the visual space to form unseen classifiers. It doesn't consider whether the two spaces are consistent or not since the semantic information is obtained independently of the visual exemplars. While our approach focuses on aligning the visual-semantic structure and then shares the aligned structures to form unseen-class prototypes in different spaces. Moreover, \cite{Changpinyo2016SynthesizedCF} learns visual classifiers independently of the semantic information while our approach automatically learns the class prototypes in the visual space by jointly leveraging the semantic information. Furthermore, to make the model more suitable to the unseen classes to tackle the challenging domain shift problem, which is not addressed in \cite{Changpinyo2016SynthesizedCF}, we propose to utilize the unseen-class semantics to make domain adaptation. Another work \cite{Wang2016LearningD} also uses structure constraints to learn visual-semantic embeddings. However, it deals with the sample structure, where the distances among samples are preserved. While our approach aligns the class structures, which aims to learn more robust class prototypes.

\section{Experiments}


\subsection{Datasets and Settings}

\textbf{Datasets}. \quad Following the new data splits proposed by \cite{Xian2017ZeroShotL}, we perform experiments on four bench-mark \textbf{ZSL} datasets, \emph{i.e.} aPascal \& aYahoo (aPY) \cite{farhadi2009describing}, Animals with Attributes (AwA) \cite{lampert2009learning}, Caltech-UCSD Birds-200-2011 (CUB) \cite{wah2011caltech}, SUN Attribute (SUNA) \cite{patterson2014sun}, to verify the effectiveness of the proposed framework. The statistics of all datasets are shown in Table \ref{table:database}.
\begin{table}[t]
\begin{small}
\begin{center}
\caption{
Statistics for attribute datasets: aPY , AwA , CUB and SUNA in terms of image numbers (\emph{Img}), attribute numbers (\emph{Attr}), training + validation seen class numbers (\emph{Seen}) and unseen class numbers (\emph{Unseen})
}
\label{table:database}
\begin{tabular}{lllll}
\hline
\noalign{\smallskip}
\textbf{Dataset}     &   \quad \emph{Img}    &  \quad \emph{Attr}   &  \quad \emph{Seen}  &  \quad \emph{Unseen} \\
\noalign{\smallskip}
\hline
\hline
\textbf{aPY} \cite{farhadi2009describing}       &  \quad 15,339   &  \quad 64   &  \quad 15 + 5     &  \quad 12   \\
\textbf{AwA} \cite{lampert2009learning}         &  \quad 30,475   &  \quad 85   &  \quad 27 + 13    &  \quad 10   \\
\textbf{CUB} \cite{wah2011caltech}              &  \quad 11,788   &  \quad 312  &  \quad 100 + 50   &  \quad 50   \\
\textbf{SUNA} \cite{patterson2014sun}           &  \quad 14,340   &  \quad 102  &  \quad 580 + 65   &  \quad 72 \\
\hline
\end{tabular}
\end{center}
\end{small}
\end{table}

\noindent \textbf{Settings}. \quad To make fair comparisons, we use the class semantics and image features provided by \cite{Xian2017ZeroShotL}. Specifically, the attribute vectors are utilized as the class semantics and the image features are extracted by the 101-layered ResNet \cite{Kaiming2016}. Parameters ($\lambda, \alpha, \beta, \gamma$) in the proposed framework are fine-tuned in the range $[0.001, 0.01, 0.1, 1, 10]$ using the train and validation splits provided by \cite{Xian2017ZeroShotL}. More details about the parameters can be seen in the supplementary material. We use the average per-class top-1 accuracy to measure the performance of our models.

\subsection{Evaluations of Different Spaces}
\label{sec:space}
\begin{figure}[t]
\centering
\includegraphics[height=3.5cm]{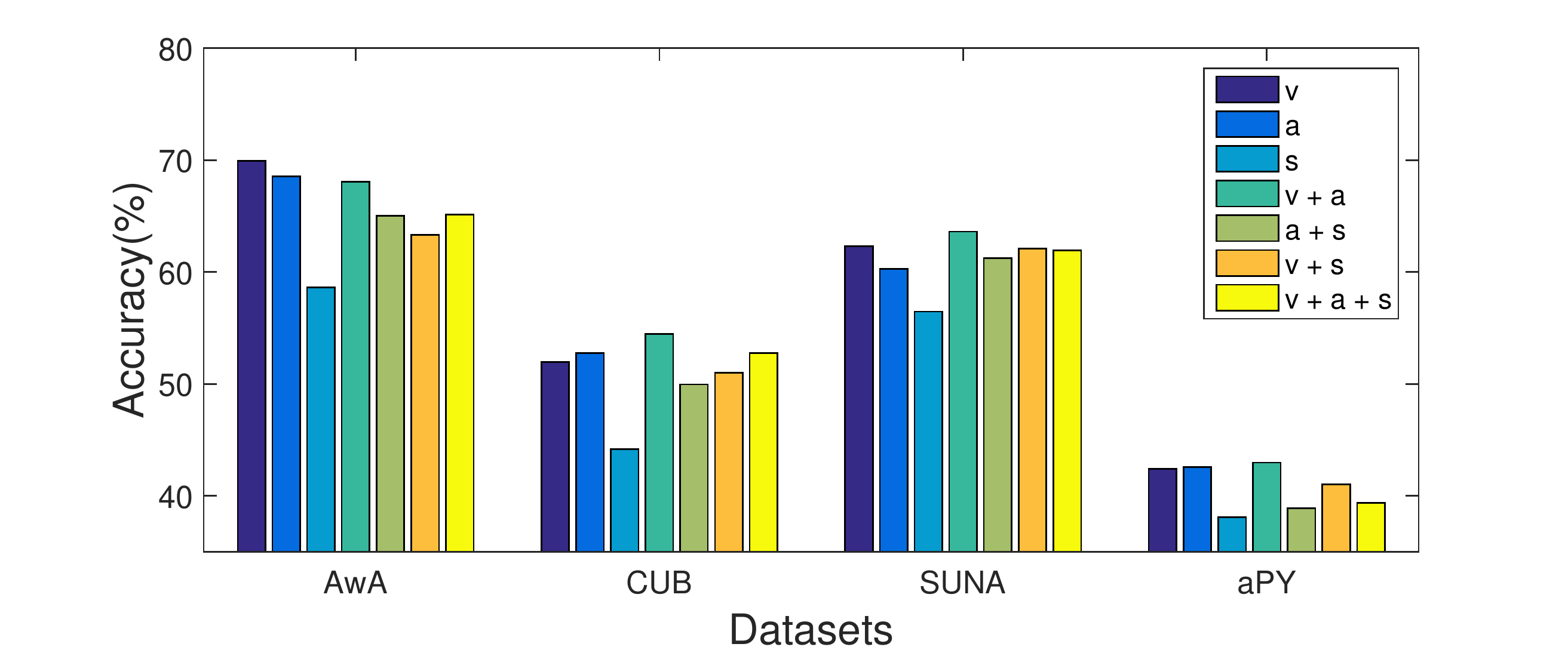}
\caption{Zero-shot recognition results via different evaluation spaces, \emph{i.e.} visual space (v), aligned space (a), semantic space (s), combination of visual space and aligned space (v + a) and other combinations, as is described in Section \ref{sec:recognition}.}
\label{fig:space}
\end{figure}

The proposed framework involves three spaces, \emph{i.e.} visual space (v), aligned space (a) and semantic space (s). As is described above, zero-shot recognition can be performed in each space independently or in the combined space, and the recognition results are shown in Figure \ref{fig:space}. It can be seen that the performance in the visual space is higher than that in the semantic space, which indicates that the incomplete semantic information is usually less discriminative. By aligning the visual-semantic structures, the discriminative property of the semantic space improves a lot, which can be inferred from the comparisons between the aligned space and the semantic space. Moreover, the recognition performance will be further improved by combining the visual space and the aligned space, since the visual space is more discriminative and the aligned space is more extensive. For AwA, the best performance is obtained in the visual space. Perhaps the visual space is discriminative enough and it is not complementary with other spaces, so combining it with others will pull down its performance.

\subsection{Comparison with State-of-the-Art}

To demonstrate the effectiveness of the proposed framework, we compare our method with several popular approaches and the recognition results on the four datasets are shown in Table \ref{table:compare}. We report our results in the best space for each dataset, as is analyzed in Section \ref{sec:space}. It can be seen that our framework achieves the best performance on three datasets and is comparable to the best approach on CUB, which indicates the effectiveness of our framework. SAE \cite{Kodirov2017SemanticAF} gets poor performance on aPY probably due to that it is not robust to the weak relations between seen and unseen classes. We owe the success of \textbf{CDL} to the structure alignment procedure. Different from other approaches, where fixed semantic information is utilized to perform the recognition task, we automatically adjust the semantic space by aligning the visual-semantic structures. Since the visual space is more discriminative and the semantic space is more extensive, it will benefit each other by aligning the structures for the two spaces. Compared with \cite{Changpinyo2016SynthesizedCF}, we get slightly lower result on CUB and this may be caused by the less discriminative class structures. CUB is a fine-grained dataset, where most classes are very similar, so less discriminative class relations could be obtained in the visual space. While \cite{Changpinyo2016SynthesizedCF} learns more complicated image classifiers to enhance the discriminative property in the visual space.
\begin{table}[t]
\begin{small}
\begin{center}
\caption{
Zero-shot recognition results on aPY, AwA, CUB and SUNA (\%)
}
\label{table:compare}
\begin{tabular}{lllll}
\hline
\noalign{\smallskip}
\textbf{Method}     &   \quad \textbf{aPY}    &  \quad \textbf{AwA}   &  \quad \textbf{CUB}  &  \quad \textbf{SUNA} \\
\noalign{\smallskip}
\hline
\hline
DAP \cite{lampert2009learning}              &  \quad 33.8   &  \quad 44.1   &  \quad 40.0   &  \quad 39.9   \\
IAP \cite{lampert2009learning}              &  \quad 36.6   &  \quad 35.9   &  \quad 24.0   &  \quad 19.4   \\
CONSE \cite{Norouzi2014ZeroShotLB}          &  \quad 26.9   &  \quad 45.6   &  \quad 34.3   &  \quad 38.8   \\
CMT \cite{Socher2013ZeroShotLT}             &  \quad 28.0   &  \quad 39.5   &  \quad 34.6   &  \quad 39.9 \\
SSE \cite{Zhang2015ZeroShotLV}              &  \quad 34.0   &  \quad 60.1   &  \quad 43.9   &  \quad 51.5   \\
LATEM \cite{Xian2016LatentEF}               &  \quad 35.2   &  \quad 55.1   &  \quad 49.3   &  \quad 55.3   \\
ALE \cite{akata2013label}                   &  \quad 39.7   &  \quad 59.9   &  \quad 54.9   &  \quad 58.1   \\
DEVISE \cite{frome2013devise}               &  \quad 39.8   &  \quad 54.2   &  \quad 52.0   &  \quad 56.5 \\
SJE \cite{akata2015evaluation}              &  \quad 32.9   &  \quad 65.6   &  \quad 53.9   &  \quad 53.7   \\
EZSL \cite{romera2015embarrassingly}        &  \quad 38.3   &  \quad 58.2   &  \quad 53.9   &  \quad 54.5   \\
SYNC \cite{Changpinyo2016SynthesizedCF}     &  \quad 23.9   &  \quad 54.0   &  \quad \textbf{55.6}   &  \quad 56.3   \\
SAE \cite{Kodirov2017SemanticAF}            &  \quad 8.3    &  \quad 53.0   &  \quad 33.3   &  \quad 40.3 \\
\hline
\textbf{CDL}(Ours)                          &  \quad \textbf{43.0}   &  \quad \textbf{69.9}   &  \quad 54.5   &  \quad \textbf{63.6} \\
\hline
\end{tabular}
\end{center}
\end{small}
\end{table}

\subsection{Effectiveness of the Proposed Framework}

\begin{figure}[t]
\centering
\includegraphics[height=3.5cm]{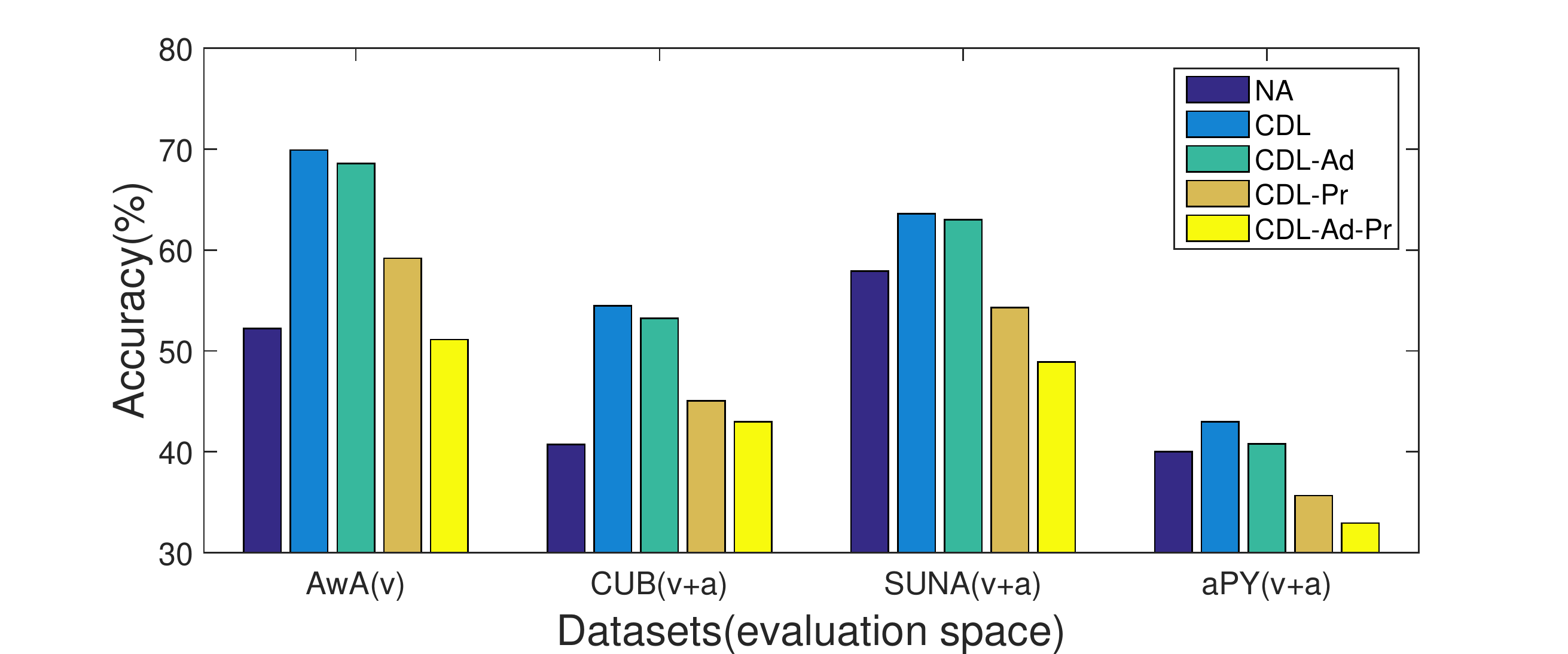}
\caption{Comparisons of different baseline methods. NA: not aligning the visual-semantic structure, as is done in the initialization period. CDL: The proposed framework. CDL-Ad: CDL without the adaptation term (second term). CDL-Pr: CDL without the prototype learning term (third term), where $P_s$ is fixed as the means of visual samples in each class. CDL-Ad-Pr: CDL without the adaptation term and the prototype learning term.}
\label{fig:terms}
\end{figure}

In order to demonstrate the effectiveness of each component proposed in our framework, we compare our approach with different submodels. The recognition task is performed in the best space according to the datasets. Specifically, for CUB, SUNA, aPY, we evaluate the performance by combining the visual space and the aligned space; for AwA, we evaluate the performance in the visual space. Figure \ref{fig:terms} shows the zero-shot recognition results of different submodels.
By comparing the performance of ``NA" and ``CDL", we can figure out that the models will improve a lot by aligning the visual-semantic structures and the less discriminative semantic space will be improved with the help of discriminative visual space. However, if the seen-class prototypes are fixed, it becomes difficult to align the structures between the two spaces and the models degrade seriously, which can be seen through the comparisons of ``CDL" and ``CDL-Pr". Moreover, the models will be more suitable to the unseen classes by utilizing the unseen-class semantic information to adapt the learning procedure, which is indicated by the comparisons of ``CDL" and ``CDL-Ad".

\subsection{Visualization of the Class Structures}

\begin{figure}[t]
\centering
\includegraphics[height=5.3cm, trim= 10 50 0 0]{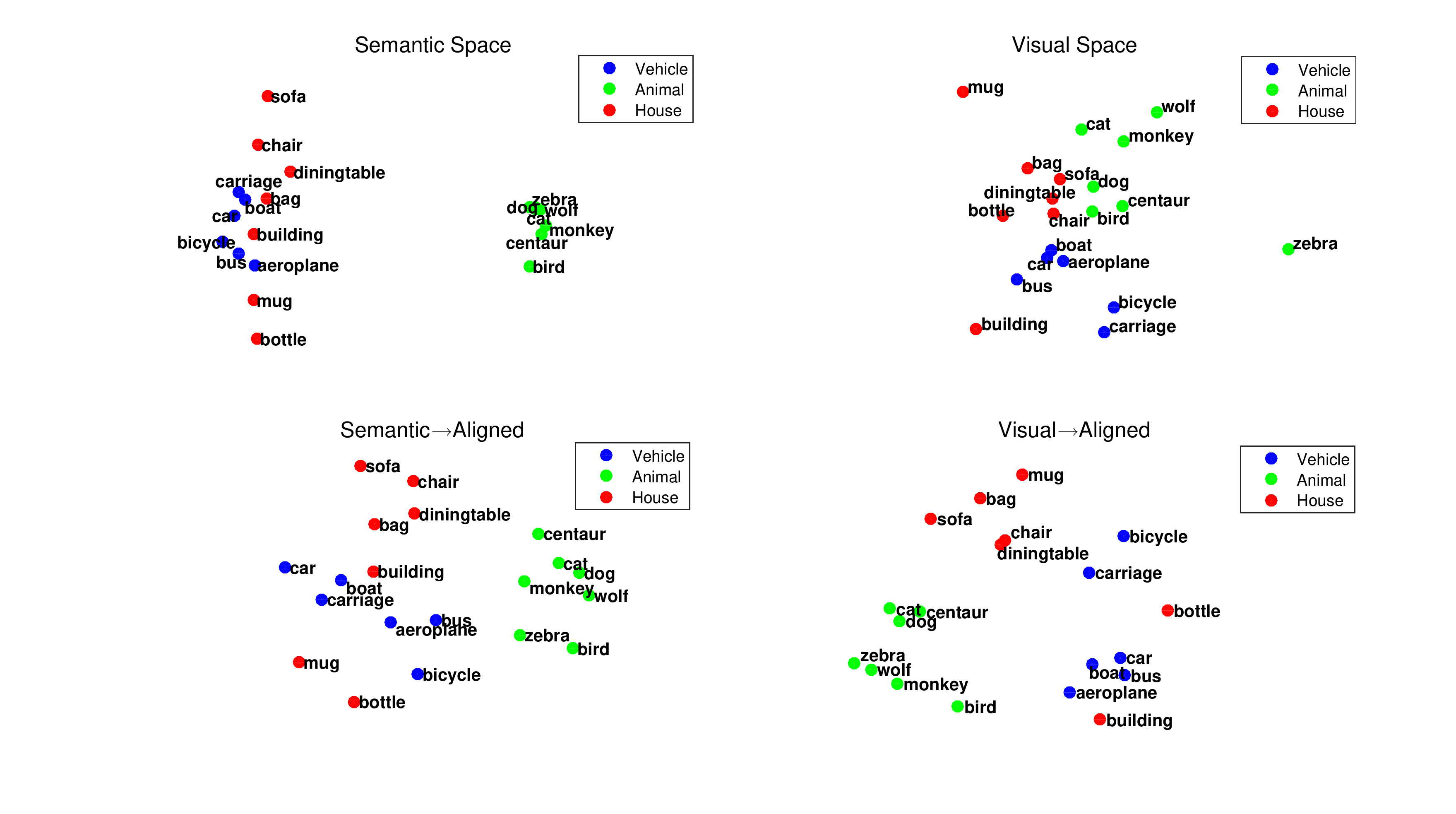}
\caption{ Visualization of the seen-class prototypes in the semantic space and visual space before and after structure alignment on aPY. To make it intuitive, the classes are manually clustered into three groups, \emph{i.e.}Vehicle, Animal and House.}
\label{fig:align}
\end{figure}

In order to have an intuitive understanding of structure alignment, we visualize the class prototypes in the visual space and semantic space on aPY, since the classes in aPY are more easy to understand. In the visual space, we obtain the class prototypes by the mean feature vector of all samples belonging to each class. In the semantic space, we get the class prototypes directly from the semantic representations. Then we use multidimensional scaling (MDS) approach \cite{Kruskal2005MultidimensionalSB} to visualize the class prototypes, where the relations of all classes are preserved. The original class structures in the semantic space and the visual space are shown in the first row of Figure \ref{fig:align}.
To make the figure more intuitive, we manually gathered the classes into three groups, \emph{i.e.} Vehicle, Animal and House. We can figure out that the class structures in the semantic space are not discriminative enough, as can be seen by the tight structures among animals, while those in the visual space are more discriminative. Moreover, the structures between these two space are seriously inconsistent, so directly sharing the structures from the semantic space to the visual space to synthesize the unseen-class prototypes will degrade the model. Therefore, we propose to learn the representation bases in each space to reformulate the class prototypes and align the class structures in a common space. It can be seen that the semantic structures become more discriminative after structure alignment. For example, in the original semantic space, \emph{dog} and \emph{cat} are mostly overlapped and they are separated after structure alignment with the help of their relations in the visual space. Thus the aligned semantic space becomes more discriminative to different classes. Moreover, the aligned structures in the two spaces become more consistent than those in the original spaces.

\subsection{Visualization of Class Prototypes}

\begin{figure}[t]
\centering
\includegraphics[height=4cm, trim= 0 50 0 0]{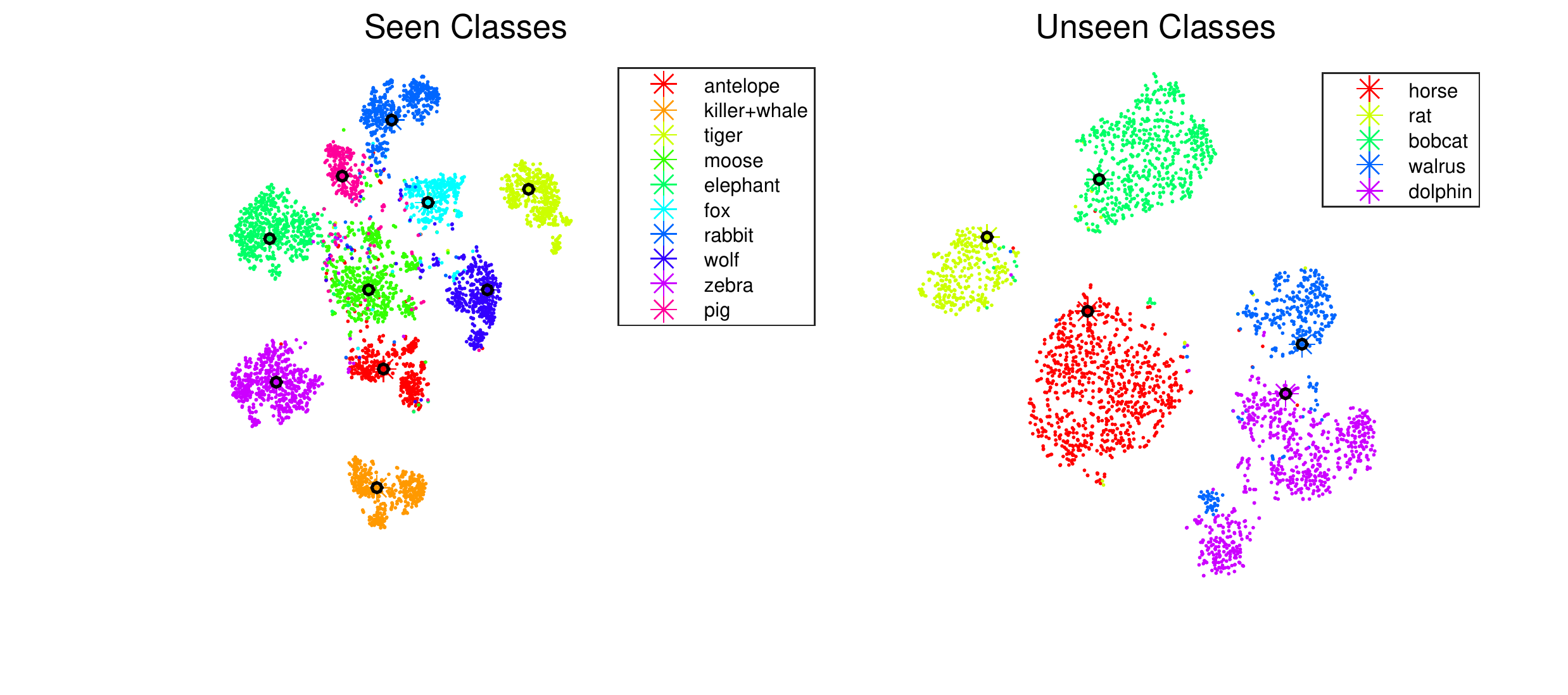}
\caption{Visualization of class prototypes on AwA in the feature space by t-SNE. The prototypes are represented by ``*" with colors corresponding to the classes. To make them visible, we use black circles to mark them.}
\label{fig:proto}
\end{figure}

The prototype of one class should locate near the samples belonging to the corresponding class. In order to check whether the prototypes are properly learned, we visualize the prototypes and corresponding samples in the visual space. To have more intuitive understanding, we choose 10 seen classes and 5 unseen classes from AwA. Then we use t-SNE \cite{Maaten2008VisualizingDU} to project the visual samples and class prototypes to a 2-D plane. The visualization results are shown in Figure \ref{fig:proto}. It can be seen that most prototypes locate near the samples belonging to the same classes. Although the unseen prototypes deviate from the centers of corresponding samples due to the fact that no corresponding images are provided for training, they are still discriminative enough to classify different classes, which shows the expansibility of our structure alignment approach for prototype learning. More visualization results can be seen in the supplementary material.

\subsection{Generalized Zero-Shot Learning}
To demonstrate the effectiveness of the proposed framework, we also apply our method to the generalized zero-shot learning (\textbf{GZSL}) task, where the seen class are also considered in the test procedure. The task for \textbf{GZSL} is to learn images classifiers $f_{gzsl}: \mathcal{X} \rightarrow \mathcal{Y}^{s} \bigcup \mathcal{Y}^{u}$. We adopt the data splits provided by \cite{Xian2017ZeroShotL} and compare our method with several popular approaches. Table \ref{table:compare2} shows the generalized zero-shot recognition results on the four datasets. It can be seen that most approaches get low accuracy on the unseen-class samples because of overfitting the seen classes, while our framework gets better results on the unseen classes and achieves more balanced results between the seen and unseen classes. By jointly aligning the visual-semantic structures and utilizing the semantic information of unseen classes to make an adaption, our model has less tendency to overfit the seen classes.
\begin{table}[t]
\scriptsize
\begin{center}
\caption{
Generalized zero-shot learning results on aPY, AwA, CUB and SUNA. ts = Top-1 accuracy of the test unseen-class samples, tr = Top-1 accuracy of the test seen-class samples, H = harmonic mean (CMT*: CMT with novelty detection). We measure top-1 accuracy in \%.
}
\label{table:compare2}
\begin{tabular}{p{2cm}|p{0.7cm}p{0.7cm}p{0.7cm}|p{0.7cm}p{0.7cm}p{0.7cm}|p{0.7cm}p{0.7cm}p{0.7cm}|p{0.7cm}p{0.7cm}p{0.7cm}}
\hline
\multirow{2}{*}{Method}&
    \multicolumn{3}{c|}{aPY}&\multicolumn{3}{c|}{AwA}&\multicolumn{3}{c|}{CUB}&\multicolumn{3}{c}{SUNA}\cr
    &ts & tr & H & ts & tr & H & ts & tr & H & ts & tr & H\cr
\hline
DAP \cite{lampert2009learning}              &4.8   &78.3   &9.0   &0.0   &\textbf{88.7}   &0.0    &1.7    &67.9   &3.3    &4.2    &25.1   &7.2 \\
IAP \cite{lampert2009learning}              &5.7   &65.6   &10.4  &2.1   &78.2   &4.1    &0.2    &72.8   &0.4    &1.0    &37.8   &1.8 \\
CONSE \cite{Norouzi2014ZeroShotLB}          &0.0   &\textbf{91.2}   &0.0   &0.4   &88.6   &0.8    &1.6    &72.2   &3.1    &6.8    &39.9   &11.6 \\
CMT \cite{Socher2013ZeroShotLT}             &1.4   &85.2   &2.8   &0.9   &87.6   &1.8    &7.2    &49.8   &12.6   &8.1    &21.8   &11.8 \\
CMT* \cite{Socher2013ZeroShotLT}            &10.9  &74.2   &19.0  &8.4   &86.9   &15.3   &4.7    &60.1   &8.7    &8.7    &28.0   &13.3 \\
SSE \cite{Zhang2015ZeroShotLV}              &0.2   &78.9   &0.4   &7.0   &80.5   &12.9   &8.5    &46.9   &14.4   &2.1    &36.4   &4.0 \\
LATEM \cite{Xian2016LatentEF}               &0.1   &73.0   &0.2   &7.3   &71.7   &13.3   &15.2   &57.3   &24.0   &14.7   &28.8   &19.5 \\
ALE \cite{akata2013label}                   &4.6   &73.7   &8.7   &16.8  &76.1   &27.5   &23.7   &62.8   &\textbf{34.4}   &\textbf{21.8}   &33.1   &26.3 \\
DEVISE \cite{frome2013devise}               &4.9   &76.9   &9.2   &13.4  &68.7   &22.4   &\textbf{23.8}   &53.0   &32.8   &16.9   &27.4   &20.9 \\
SJE \cite{akata2015evaluation}              &3.7   &55.7   &6.9   &11.3  &74.6   &19.6   &23.5   &59.2   &33.6   &14.1   &30.5   &19.8 \\
EZSL \cite{romera2015embarrassingly}        &2.4   &70.1   &4.6   &6.6   &75.6   &12.1   &12.6   &63.8   &21.0   &11.0   &27.9   &15.8 \\
SYNC \cite{Changpinyo2016SynthesizedCF}     &7.4   &66.3   &13.3  &8.9   &87.3   &16.2   &11.5   &\textbf{70.9}   &19.8   &7.9    &\textbf{43.3}   &13.4 \\
SAE \cite{Kodirov2017SemanticAF}            &0.4   &80.9   &0.9   &1.8   &77.1   &3.5    &7.8    &54.0   &13.6   &8.8    &18.0   &11.8 \\
\hline
\textbf{CDL}(Ours)       &\textbf{19.8}   &48.6   &\textbf{28.1}   &\textbf{28.1}   &73.5   &\textbf{40.6}   &23.5   &55.2   &32.9   &21.5   &34.7   &\textbf{26.5} \\
\hline
\end{tabular}
\end{center}
\end{table}

\section{Conclusions}

In this paper, we propose a coupled dictionary learning framework to align the visual-semantic structures for zero-shot learning, where unseen-class prototypes are learned by sharing the aligned structures. Extensive experiments on four bench-mark datasets show the effectiveness of the proposed approach. The success of \textbf{CDL} should be owing to three characters. First, instead of using the fixed semantic information to perform recognition task, our structure alignment approach shares the discriminative property lying in the visual space and the extensive property lying in the semantic space, which benefits each other and improves the incomplete semantic space. Second, by utilizing the unseen-class semantics to adapt the learning procedure, our model is more suitable for the unseen classes. Third, the class prototypes are automatically learned by sharing the aligned structures, which makes it possible to directly perform recognition task using simple nearest neighbour approach. Moreover, we combine the information of multiple spaces to improve the recognition performance.

\textbf{Acknowledgements.} \quad This work is partially supported by Natural Science Foundation of China under contracts Nos. 61390511, 61772500, 973 Program under contract No. 2015CB351802, Frontier Science Key Research Project CAS No. QYZDJ-SSW-JSC009, and Youth Innovation Promotion Association CAS No. 2015085.

\bibliographystyle{splncs}
\bibliography{egbib}
\end{document}